\documentclass[a4paper,fleqn]{cas-dc}

\usepackage[numbers]{natbib}
\usepackage{amssymb}
\usepackage{amsmath}
\usepackage{color}
\usepackage{graphicx}
\usepackage{tabularx}
\usepackage{multirow}
\usepackage{diagbox}
\usepackage{booktabs}
\usepackage{footnote}
\usepackage{threeparttable}
\usepackage{lineno}

\begin{document}

\shorttitle{A Multi-View Framework for BGP Anomaly Detection via GAT}    

\shortauthors{Songtao Peng \emph{et al.}}  

\title [mode = title]{A Multi-View Framework for BGP Anomaly Detection via Graph Attention Network}



%
\author[1]{Songtao Peng}

\author[1]{Jiaqi Nie}

\author[1,2]{Xincheng Shu}
\cormark[1]

\author[1]{Zhongyuan Ruan}[orcid=0000-0003-4953-070X]
\cormark[1]
    
\author[1]{Lei Wang}

\author[3]{Yunxuan Sheng}

\author[1]{Qi Xuan}

\address[1]{Institute of Cyberspace Security, Zhejiang University of Technology,  Hangzhou 310023, China}
\address[2]{Department of Electrical Engineering, City University of Hong Kong, Hong Kong 999077, China}          
\address[3]{School of Computer Science, University of Nottingham Ningbo China, Ningbo 315100, China}

\cortext[cor1]{Corresponding author}
\cortext[1]{E-mail addresses: zyruan@zjut.edu.cn}
\cortext[1]{E-mail addresses: sxc.shuxincheng@foxmail.com}

\fntext[fn1]{S.Peng developed this research and conducted experiments. J.Nie drew experimental diagrams. X.Shu developed the methodology. L.Wang collected and processed data. S.Peng, J.Nie, X.Shu, Z.Ruan, Y.Sheng and Q.Xuan wrote the manuscript together.}


\begin{abstract}
As the default protocol for exchanging routing reachability information on the Internet, the abnormal behavior in traffic of Border Gateway Protocols (BGP) is closely related to Internet anomaly events. The BGP anomalous detection model ensures stable routing services on the Internet through its real-time monitoring and alerting capabilities. Previous studies either focused on the feature selection problem or the memory characteristic in data, while ignoring the relationship between features and the precise time correlation in feature (whether it's long or short term dependence). In this paper, we propose a multi-view model for capturing anomalous behaviors from BGP update traffic, in which Seasonal and Trend decomposition using Loess (STL) method is used to reduce the noise in the original time-series data, and Graph Attention Network (GAT) is used to discover feature relationships and time correlations in feature, respectively. Our results outperform the state-of-the-art methods at the anomaly detection task, with the average F1 score up to 96.3\% and 93.2\% on the balanced and imbalanced datasets respectively. Meanwhile, our model can be extended to classify multiple anomalous and to detect unknown events.
\end{abstract}

\maketitle

\section{Introduction}
\begin{figure}[htp]
    \centering
    \includegraphics[width=8.0cm]{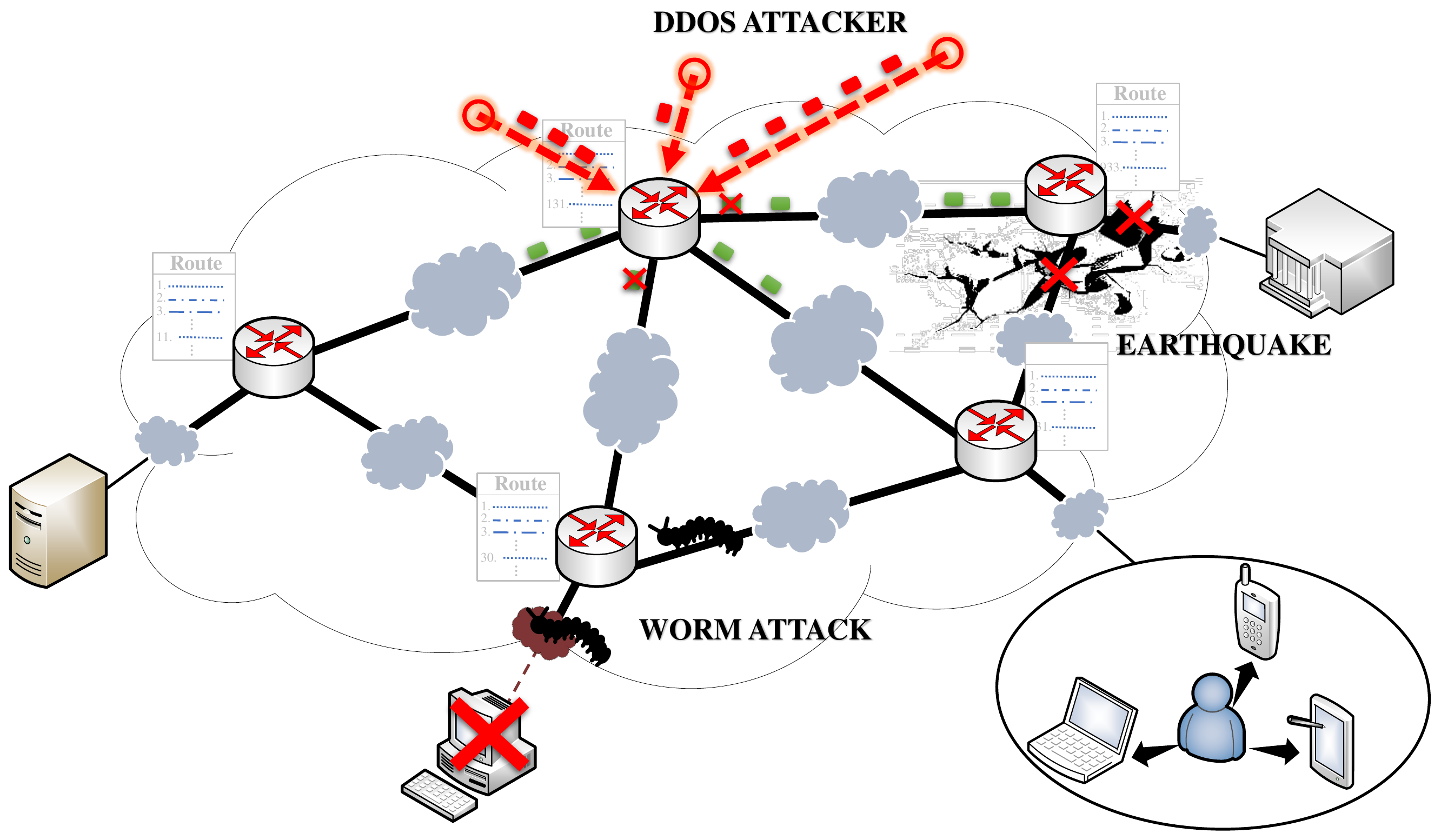}
    \caption{Cyber security events in the Internet.}
    \label{fig:route_network}
    \vspace*{-0mm}
\end{figure}

While the growth of the Internet promotes many fields, many issues have emerged that affect network security and stability. The BGP, as an inter-domain routing communication protocol, is responsible for managing Network Reachable Information (NRI) between Autonomous Systems (ASes), ensuring global reachability of information \cite{rekhter1994border}. However, BGP is vulnerable to hijacking, misconfiguration, DDoS attacks, and natural disasters (As shown in Fig. \ref{fig:route_network}). Statistics show that approximately 20\% of the hijacking and misconfigurations lasted less than 10 minutes but were able to pollute 90\% of the Internet in less than 2 minutes \cite{shi2012detecting}. For instance, on 24 February 2008, Pakistan Telecom (AS17557) published the prefix 208.65.153.0/24 without authorization, resulting in YouTube traffic being hijacked worldwide \cite{cheng2016ms}. Statistical or accessibility-based methods \cite{zhang2008ispy, theodoridis2013novel} for monitoring and determining nominal ranges are limited by the need for extensive expert knowledge and human resources \cite{hundman2018detecting}. Since hazards can not be identified until the contamination of a portion of detectable features of the ASes reaches the threshold, most methods are time-lagged \cite{moriano2021using}. So it is of great importance to accurately detect anomaly information or behaviors in BGP traffic while the network is flooded with uncertainties.

The current machine learning models on BGP anomaly detection are based on the dataset constructed from the records of BGP update traffic packets, which are partitioned into "anomalous" and "normal" samples. To improve the anomaly detection performance, these models mainly focus on the following two dimensions \cite{wen2020time}: feature dimension (to choose appropriate features) and time dimension (to choose appropriate models). For traditional machine learning methods (like SVM), the data (i.e., the features) at different timestamps are considered to be independent samples \cite{de2011anomaly,al2012machine,li2014classification} in which the time correlations are totally ignored. To overcome this shortcoming, the neural network based methods take into account the memory effect to enhance the detection performance. Despite these efforts, the factors such as the relationship between features and the precise time correlations for each feature are largely ignored, which may have significant effects on the anomaly detection task.

In this paper, we propose a unified framework that considers the ignored factors mentioned above. To augment the data, based on the window slicing, we incorporate STL decomposition and construct two parallel pipelines in feature and time dimensions by using GAT to capture feature relationships and temporal dependencies in time series data. Through extensive experiments, we verify the effectiveness of our model. The contributions of our paper are summarized as follows:
\begin{itemize}
  \item We propose a novelty framework for BGP anomaly detection at multi-view which applies STL decomposition to weaken noisy data and to enhance the area difference between positive and negative samples, and focus on feature relationships and temporal correlation in features for the first time.
  \item We extend the BGP anomalous events to six pieces for a total of 12 different datasets, conduct experiments from the perspective of balanced and imbalanced samples, respectively.
  \item Through a series of ablation experiments, we demonstrate the practicality of data augmentation methods (window slicing and STL) and multiple perspectives (feature-based GAT and temporal-based GAT). We conduct interpretable experiments on our model and extend it to classify multiple anomalous events and to detect unknown events.
\end{itemize}

The rest of paper is organized as follows. Section 2 introduces related work. Section 3 details the design and implementation of our framework. We have conducted extensive experiments and corresponding analyses in section 4. Section 5 concludes this work.

\section{Related Work}
In this section, we describe BGP anomalies and the ML-based detection models proposed in recent years.
\subsection{BGP Anomaly Detection}
The abnormal behaviors of BGP and its serious consequences draw the attention of every network operator. BGP anomaly refers to harmful changes of BGP behavior that may cause thousands of anomalous BGP updates. A single BGP update is classified as an anomaly if it contains an invalid AS number, invalid or reserved IP prefixes, AS-PATH without a physical equivalent, etc \cite{wubbeling2014inter}. In addition, a set of BGP updates can also be classified as an anomaly if the characteristics show a rapid change in the number of BGP updates, or contain longest and shortest paths, etc \cite{al2015detecting}. The studies of BGP anomaly detection can help network operators to protect their networks \cite{al2016bgp}.

\subsection{ML-based BGP Anomaly Detection}
Machine learning-based BGP anomaly detection methods in the past two decades can be roughly divided into two types from two different enhancement perspectives: feature dimension (to choose appropriate features) and time dimension (to choose appropriate models).

\textbf{Feature dimension:} The quantity or quality of features directly affects the performance of anomaly detection. Testart \emph{et al.} \cite{testart2019profiling} extract information from network operator mailing lists and train a machine learning model to automatically identify Autonomous Systems (ASes) that exhibit characteristics similar to serial hijackers. In addition, Urbina Cazenave \emph{et al.} \cite{de2011anomaly} show the SVM performs better than decision trees and Naive Bayes while using data mining algorithms to classify BGP events. Trajkovic \emph{et al.} \cite{al2012machine} select 10 features by Fisher and minimum Redundancy Maximum Relevance (mRMR) scoring algorithms to achieve significant classification performance. In recent two years, Arai \emph{et al.} \cite{arai2019selection} use the popular feature selection algorithms, wrapper as well as several filter-based algorithms for feature ranking, and Xu \emph{et al.} \cite{xu2020bgp} apply neural networks to automatically extract features from large-scale data to achieve anomaly classification. Innovatively, Sanchez \emph{et al.} \cite{sanchez2019comparing} identify graph features to detect BGP anomalies, which are arguably more robust than traditional features. Subsequently, Hoarau \emph{et al.} \cite{hoarau2021suitability} are also further evaluated and compared the accuracy of machine learning models using graph features and statistical features on both large and small scale BGP anomalies. Both of these provide theoretical support for the BGP anomaly detection methods using graphs.

\textbf{Time dimension:} Sequential models take a further step in anomaly classification. Ding \emph{et al.} \cite{ding2016detecting} use the feature selection algorithm (mRMR) to select individual features and compares the effects of classification between SVM and LSTM. Chauhan \emph{et al.} \cite{chauhan2015anomaly} propose stacked LSTM network models to learn higher-level temporal patterns for anomaly detection. Cheng \emph{et al.} \cite{cheng2016ms} use LSTM models to identify worm attacks such as Nimda and Code Red I. And they propose the MSLSTM model in the following work \cite{cheng2018multi} which uses a discrete wavelet transform to obtain multi-scale temporal information, combining attention to construct a two-layer LSTM architecture. To reduce the impact of inaccurate supervision, Dong \emph{et al.} \cite{dong2021isp} design a self-attention-based LSTM model to self-adaptively mine the differences in BGP anomaly categories.

Despite the great progress, there remain many issues to address. For example, the correlation of added feature with the target is low, while the proportion of this feature is quite high, and the performance of sequential models fluctuates significantly across different anomalous events. To overcome the above problems, in this paper, we propose a new framework for the BGP anomaly detection task, and further apply this method to classify unknown events.

\section{Methodology}
In this section, we define the problem that needs to be discussed in detail and design our model with the idea as: focus on the multi-perspective features of data through data augmentation and on the multi-view relational characteristics of events through feature relationships and temporal dependencies. The overall framework of the model is shown in Figure \ref{fig:Framework}, and is designed as follows:
\begin{itemize}
 \item [1)] 
  The methods such as window slicing and STL decomposition are applied to enhance the original data. After this, a new sample with shape $m\times5k$ ($m$ is the number of the original samples in the window and $k$ is the number of the original sample features) is obtained.
 \item [2)]
  We propose the M-GAT structure, which applies feature-based and temporal-based GAT to process data, focusing on the feature correlation and temporal dependency of the data. Note that the shape of the output matrix of GAT layer is also $m\times 5k$.
 \item[3)]
  To retain the important information of the output  of  each  layer  without  causing  redundancy, we stitch the multi-channel data with specified weights after the GAT layer. Then we use LSTM to get the final prediction result $\boldsymbol{\hat{y}}$.
\end{itemize}

\subsection{Problem Definition}
We introduce a classification task on BGP anomaly detection to analyze Internet anomalous events. We consider this task as a multivariate time series anomaly detection, which is defined as follows: The input data is denoted by $\boldsymbol{X =\left\{x_{1}^{k}, x_{2}^{k}, \ldots, x_{n}^{k}\right\} \in \mathbb{R}^{n \times k}}$, where $n$ is the maximum length of timestamps, and $k$ is the number of features for each sample.  The task of anomaly detection is to produce an output vector $\boldsymbol{Y =\left\{y_{1}, y_{2}, \ldots, y_{n}\right\} \in \mathbb{R}^{n}}$ by the proposed method, where $y_i \in \left\{0, 1\right\}$ denotes whether the $i^{th}$ timestamp $\boldsymbol{x_{i}^{k}}$ is  anomalous.

\subsection{Data Augmentation}
\begin{figure}[htp]
    \centering
    \includegraphics[width=8.0cm]{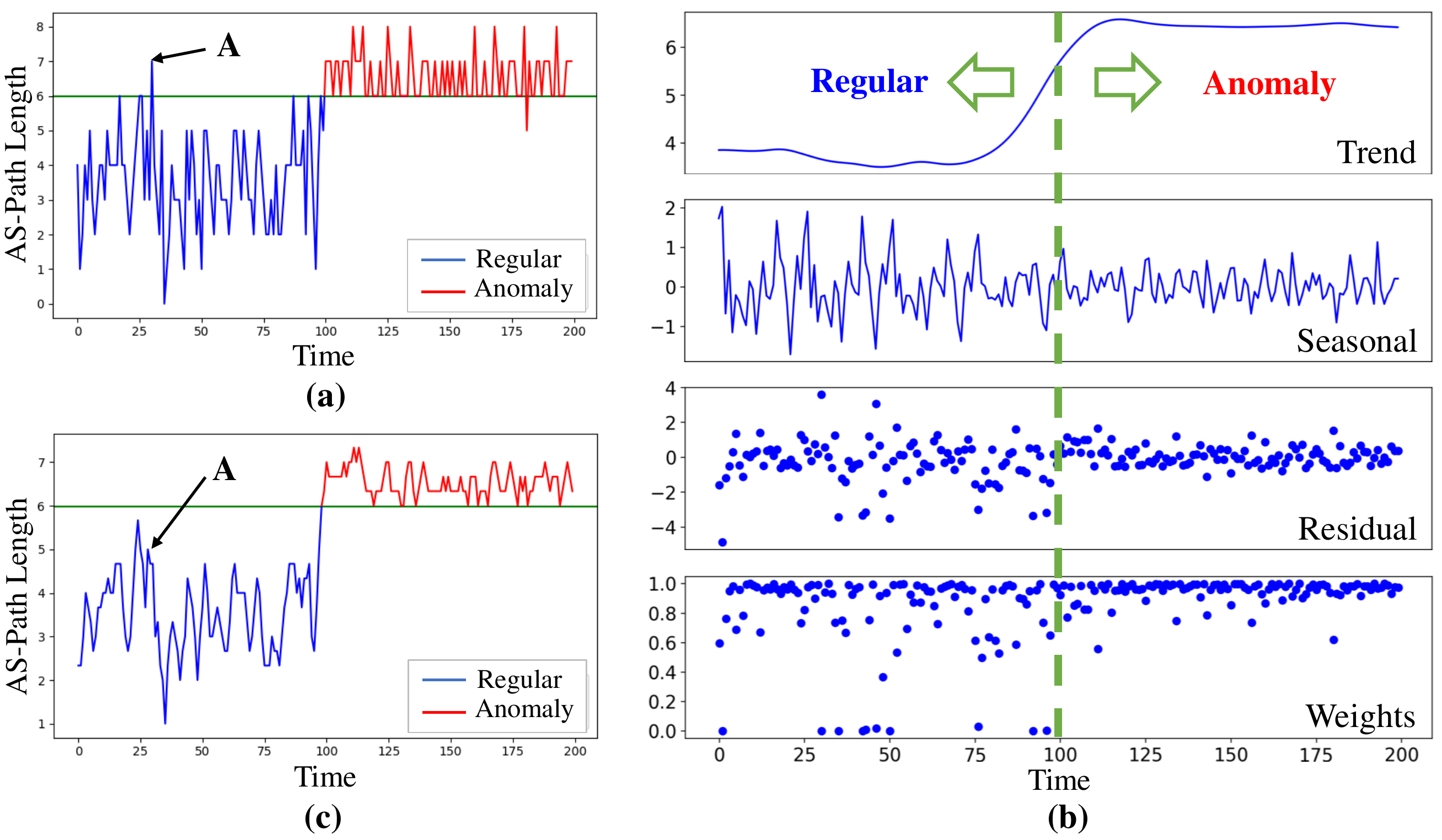}
    \caption{(a) Initial data, (b) the process of STL decomposition, and (c) result of window slicing.}
    \label{fig:STL_Window}
    \vspace*{-0mm}
\end{figure}

Machine learning based BGP anomaly detection methods \cite{de2011anomaly,al2012machine,kim2018web,fonseca2019bgp} extract the relevant feature values from the information packets, then use the classifier for anomaly detection. In real situations, a small portion of normal data will inevitably be mistaken for fluctuations or noise. As shown in Figure \ref{fig:STL_Window}(a), the curve represents 600 consecutive samples of the \emph{AS-Path\ Length} feature in the dataset \emph{Slammer}. The normal (blue curve) and abnormal (red curve) samples are separated by the green line. Take the green line as the evaluation criterion, the data point $A$ will be misclassified as an anomalous sample with a high probability. Data augmentation can reduce or even eliminate the impact of such noisy data by increasing the size and quality of the data \cite{lu2008network, wen2020time}. In our model, we use the methods such as window slicing and STL.

\begin{figure*}[htp]
    \centering
    \includegraphics[width=17cm]{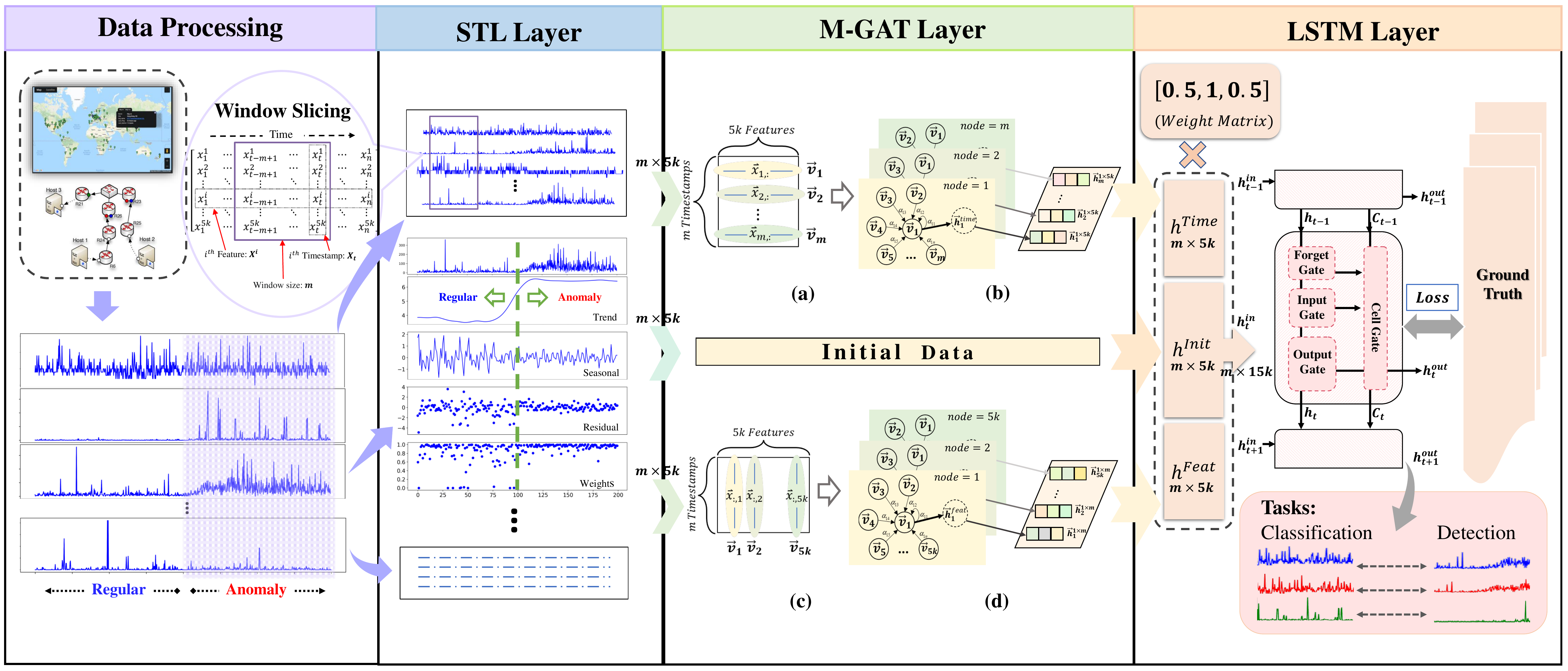}
    \caption{The framework of our model.}
    \label{fig:Framework}
    \vspace*{-0mm}
\end{figure*}

\subsubsection{Seasonal and Trend decomposition using Loess (STL)}
By using the STL \cite{robert1990stl}, a time series can be decomposed into seasonal, trend, and remainder (residual) terms. The trend term shows the direction of persistent increasing or decreasing in data, the seasonal term shows the seasonal factors over a fixed period, and the residual term means the noise of time-series. We denote the above three terms by $\boldsymbol{T_t}$, $\boldsymbol{S_t}$, and $\boldsymbol{R_t}$ respectively, and the original time series by $\boldsymbol{Y_t}$ ($t=1,...,N$). Thus we have
\begin{equation}
    \boldsymbol{Y_t = T_t + S_t + R_t}
\end{equation}
Furthermore, we introduce a robustness weight indicator $\boldsymbol{W_t}$ at each time point $t$ to reflect the extremity of the value of the residual term.

The STL decomposition is chosen for three reasons:
\begin{itemize}
  \item [1)] 
  The trend and seasonal terms are robust and do not distort the data by transient anomalous behavior.       
  \item [2)]
  The residual term and robustness weight indicator represent the anomaly-related deviations.
  \item [3)]
  STL is simple to be implemented and can quickly calculate long-term sequences.
\end{itemize}
The process of STL decomposition is shown in figure \ref{fig:STL_Window}(b). In this case, the input data can be expressed as a sequence $\boldsymbol{\left\{x_{1}^{5k}, x_{2}^{5k}, \ldots, x_{n}^{5k}\right\}}$, where $5k$ implies the information of five parts that three terms mentioned above, robustness weight indicator, and original data each with $k$ samples.

\subsubsection{Window Slicing}
Window slicing \cite{le2016data} is a subsampling method that extracts consecutive slices from the original time series, extending strongly correlated data with more feature dimensions on the original time series dataset. The length of the window slice (denoted by $m$) is an adjustable parameter. By traversing the entire initial time series in step of $1$, one can get different time windows $\boldsymbol{\left\{x_{t-m+1}^{k}, x_{t-m+2}^{k}, \ldots, x_{t}^{k}\right\}}$ via varying $t$. After normalization, each time window is denoted as $\boldsymbol{z_{t-m+1}^{m \times k}}$, where the label with the highest occurrence frequency among the $m$ samples is chosen as the new label of $\boldsymbol{z_{t-m+1}^{m \times k}}$. By averaging the previous 600 test samples, the result is shown in Figure \ref{fig:STL_Window}(c).

\subsection{Multi-view Graph Attention Network (M-GAT)}
Graph attention network (GAT) \cite{velivckovic2017graph} achieves advanced performance on many classification problems \cite{cheng2018multi,zhao2020multivariate,ding2021cross} by using the attention mechanism \cite{mnih2014recurrent}. The input of stacking graph attention layers in GAT is a set of node feature vectors, $\mathbf{h}=\boldsymbol{\left\{\vec{h}_{1}, \vec{h}_{2}, \ldots, \vec{h}_{n}\right\}, \vec{h}_{i}} \in \mathbb{R}^{F}$, where $n$ is the number of nodes, and $F$ is the number of features in each node. The final output is a new set of feature vectors with the number of nodes $n$, $\mathbf{h}^{\prime}=\boldsymbol{\left\{\vec{h}_{1}^{\prime}, \vec{h}_{2}^{\prime}, \ldots, \vec{h}_{n}^{\prime}\right\}, \vec{h}_{i}^{\prime}} \in \mathbb{R}^{F}$. The formula for obtaining $\boldsymbol{\vec{h}_{i}^{\prime}}$ is as follows:
\begin{equation}
    \boldsymbol{\vec{h}_{i}}=\sigma\left(\sum_{j \in \boldsymbol{N_{i}}} \boldsymbol{\alpha_{i j} W \vec{h}_{j}}\right)
\end{equation}
where $\boldsymbol{W}$ is the weight matrix multiplied with the features, $\boldsymbol{\sigma(\cdot)}$ is the nonlinear activation function \cite{xu2015empirical}, $\boldsymbol{N_{i}}$ represents the set of all neighbors of node $i$, and $\boldsymbol{\alpha_{i j}}$ is attention coefficient in GAT which is calculated as:
\begin{small}
\begin{equation}
    \boldsymbol{\alpha_{i j}}=\frac{\exp \left(\operatorname{LR}\left(\overrightarrow{\boldsymbol{\mathrm{a}}^{T}}\left[\boldsymbol{W \vec{h}_{i} \| W \vec{h}_{j}}\right]\right)\right)}{\sum_{k \in \boldsymbol{N_{i}}} \exp \left(\operatorname{LR}\left(\overrightarrow{\boldsymbol{\mathrm{a}}^{T}}\left[\boldsymbol{W \vec{h}_{i} \| W \vec{h}_{k}}\right]\right)\right)}
\end{equation}
\end{small}where ${(\cdot)}^T$ represents transposition and $||$ is the concatenation operation. $\overrightarrow{\boldsymbol{\mathrm{a}}} \in \mathbb{R}^{2 F}$ is the weight matrix between the connected layers. The $LR$ (i.e. $LeakyReLu$) function is applied to the output layer of the feedforward neural networks, which resets all negative numbers of the output to 0.2.

$\boldsymbol{\vec{h}_{i}^{\prime}}$ is the new feature vector of node $i$ which highlights the strong or weak relationships between adjacent nodes and the importance of a node itself. In the following sections, we will introduce \emph{feature-based} GAT and \emph{temporal-based} GAT, respectively.

\subsubsection{Feature-based GAT}
In our dataset (in which the features are mainly divided into $Volume\ Features$ and \emph{AS-path\ Features}), it is found that, for some certain features, there are large differences in values between different categories, and very similar distributions in feature values within the same category. Therefore, in the absence of a priori knowledge, it is necessary to explore the interrelationship between features.

As shown in Figure \ref{fig:Framework} (a), we construct a fully connected graph of time-series samples in each window, where each feature is taken as a node and the interrelationship between features is taken as an edge. Then we calculate the corresponding attention coefficient according to formula (2) and update the node vector. Specifically, we create a sequence for each window slice, $\boldsymbol{\left\{\vec{x}_{1:m, 1}, \vec{x}_{1:m, 2}, \ldots, \vec{x}_{1:m, 5k}\right\}}$, where $5k$ is the number of features, and $m$ is the number of samples in the window. Note that each unit in the sequence is a vector of shape $1 \times m$. Figure \ref{fig:Framework} (b) shows how the attention coefficient and the output $\boldsymbol{\vec{h}_{i}^{Feat}}$ of the specific node $i$ can be obtained through the attention mechanism. Here $\boldsymbol{\vec{v}_{i}}$ denotes $\boldsymbol{\vec{x}_{1:m,i}}, i \in(1, 5k)$. The final output is a matrix of shape $m \times 5k$.

\begin{table*}[htb]
\caption{Details of the abnormal event datasets}
\centering
\renewcommand\arraystretch{1.3} 
\begin{threeparttable}
\begin{tabular}{c c c c c c c c}
\toprule[2pt]
Event             & Total/Anomaly($AS_1$,...,$AS_n$) & Time & collector & AS Number & No.
\\ \midrule
Code Red I        & 7136 * 3 / 526, 472, 526  & 2001.07.19 - 2001.07.20 & rrc04 (Geneva) & 513, 559, 6893  & 1
\\
Nimda             & 10336 / 3535 & 2001.09.15 - 2001.09.23 & rrc04 (Geneva) & 513  & 2
\\
Slammer           & 7200 / 1130 & 2003.01.23 - 2003.01.27 & rrc04 (Geneva) & 513  & 3
\\
Moscow Blackout   & 7200 * 2 / 171, 171 & 2005.05.23 - 2005.05.27 & rrc05 (Vienna) & 1853,12793  & 4
\\
Japan Earthquake  & 7200 / 387 & 2011.03.09 - 2011.03.13 & rrc06 (Japan) & 2497 & 5
\\
Malaysian Telecom & 7200 * 4 / 103, 154, 185, 107 & 2015.06.10 - 2015.06.14 & rrc04 (Geneva) & 513,20923,25091,34781 & 6
\\
\bottomrule[2pt]
\end{tabular}
\label{tab:Datasets}
\end{threeparttable}
\vspace{-0mm}
\end{table*}

\subsubsection{Temporal-based GAT}
Considering that in real time series, previous states may have an impact on the current state, existing methods generally use models with memory capabilities (RNN, LSTM) to process the time series data. Based on this, we establish a temporal-based GAT layer.

We focus on the relationship of each pair of timestamps in the window and construct a fully connected graph of timestamps (treating as nodes). Full connectivity reflects the short-term and long-term relationships in all pairs of timestamps. As shown in Figure \ref{fig:Framework} (c), the input is time series $\boldsymbol{\left\{\vec{x}_{1,1:5k}, \vec{x}_{2,1:5k}, \ldots, \vec{x}_{m,1:5k}\right\}}$, and the output $\boldsymbol{\vec{h}_{i}^{Time}}$ is a matrix of $m \times 5k$.

\subsection{Long short-term memory (LSTM)}
We use the LSTM model \cite{hochreiter1997long} to predict the sample categories. LSTM solves the gradient disappearance and gradient explosion problems in long sequence training by replacing a single neural network layer of RNN with multiple neural network layers. It is composed of a cell, an input gate, an output gate and a forget gate. The three gates are used to control the flow of information into or out of the cell $\textbf{\emph{c}}$. The whole process can be described by the following formulas:
\begin{align}
    \boldsymbol{f_{t}}&=\boldsymbol{\sigma_{g}\left(W_{f} x_{t}+U_{f} h_{t-1}+b_{f}\right)}\\
    \boldsymbol{i_{t}}&=\boldsymbol{\sigma_{g}\left(W_{i} x_{t}+U_{i} h_{t-1}+b_{i}\right)}\\
    \boldsymbol{o_{t}}&=\boldsymbol{\sigma_{g}\left(W_{o} x_{t}+U_{o} h_{t-1}+b_{o}\right)}\\
    \boldsymbol{\tilde{c}_{t}}&=\boldsymbol{\sigma_{c}\left(W_{c} x_{t}+U_{c} h_{t-1}+b_{c}\right)}\\
    \boldsymbol{c_{t}}&=\boldsymbol{f_{t} * c_{t-1}+i_{t} * \tilde{c}_{t}} \\
    \boldsymbol{h_{t}}&=\boldsymbol{o_{t} * \sigma_{h}\left(c_{t}\right)}
\end{align}
where $\boldsymbol{W_{\alpha}, U_{\alpha}, b_{\alpha}}$ ($\alpha \in \{f, i, o, c\}$) represent trainable input weights, recurrent connections weights and bias vectors, respectively. $\boldsymbol{\sigma_g}$ is sigmoid function, $\sigma_c$ is hyperbolic tangent function, $\boldsymbol{\sigma_h}$ is hyperbolic tangent function or $\boldsymbol{\sigma_h}(x)=x$. $\boldsymbol{x_t}$ is input vector to the LSTM unit at time $t$, $\boldsymbol{f_t}$ is forget gate's activation vector, $\boldsymbol{i_t}$ is input gate’s activation vector, $\boldsymbol{o_t}$ is output gate’s activation vector, $\boldsymbol{h_t}$ is hidden state vector also know as output vector of the LSTM unit, $\boldsymbol{\tilde{c}_{t}}$ is cell input activation vector, and $\boldsymbol{c_t}$ is cell state vector.

Finally, the vector of prediction results $\boldsymbol{\hat{y}}$ can be obtained from $\boldsymbol{h_t[-1]}$. We use cross entropy as a loss function to estimate the difference between the predicted and true label values, then optimize the model and determine the parameters of each layer. The formula is as follows:
\begin{small}
\begin{equation}
    \begin{aligned}
        L(\boldsymbol{\hat{y}}, \boldsymbol{y})\!=\!\boldsymbol{W_m[y]}\left(\boldsymbol{\!-\!\hat{y}[y]}\!+\!\log \left(\sum_{j=0}^{n} e^{\boldsymbol{\hat{y}[j]}}\right)\right)
    \end{aligned}
\end{equation}
\end{small}where $\boldsymbol{y}$ is the true label, and we increase the heterogeneity of different categories by presetting the weight matrix $\boldsymbol{W_m}$. To reduce the influence of $\boldsymbol{{h}^{Feat}}$ and $\boldsymbol{{h}^{Time}}$, we set $\boldsymbol{W_m}$ as $[0.5, 1, 0.5]$.

\section{Experimental Evaluation and Discussion}
To demonstrate the effectiveness and universality of our model for the BGP anomaly detection problem, in this section, we evaluate the detection metrics for different anomalous events, the effectiveness of module and parameter selection in the model, and the performance of the model in unbalanced samples. We also evaluate the interpretability of the model and its performance on multi-classification task.

\subsection{Datasets}
BGP anomaly detection technology uses BGP update packets to detect BGP anomalies. Route Views \cite{routeviews} and RIPE NCC \cite{ripencc} are the most well-known repositories that provide free downloads of historical BGP update data. By parsing and preliminary pre-processing \cite{fonseca2019deep} of the data downloaded from the platforms, we obtain the anomaly event dataset, as shown in Table \ref{tab:Datasets}.

To avoid excessive disparity in sample size between different categories, we narrow the event scope of the data without affecting the anomalies, at the mean time, we incorporate the traffic of different AS in the same event. We extract relevant information from BGP update packets and count them at intervals of one minute to obtain several relevant feature values. In general, these features can be divided into \emph{volume features} and \emph{AS-PATH features} \cite{al2016bgp,peng2021inferring}, such as the number of Network Layer Reachability Information (NLRI) prefixes announced or withdrawn, the average AS-PATH length, etc. In total, we select 46 features \cite{fonseca2019deep} which are listed in Table \ref{tab:Features_46}.

\begin{table}[htb]
\caption{46 features from BGP update message.}
\centering
\renewcommand\arraystretch{1.05} 

\newcommand{\tabincell}[2]{\begin{tabular}{@{}#1@{}}#2\end{tabular}}
\begin{tabular}{c c c}
\toprule[2pt]
\textbf{Index} & \textbf{Definitions} & \textbf{Category}\\ \hline
1     & Number of announcements & volume\\ 
2     & Number of withdrawals & volume\\ 
3     & Number of duplicate announcements & volume\\ 
4     & 37	Number of NLRI announcements & volume\\ 
5     & \tabincell{c}{Number of non-duplicate\\ announcements} & volume\\
6     & Number of flaps & volume\\
7     & \tabincell{c}{Number of new announcements\\ after withdraw} & volume\\
8     & Number of plain new announcements & volume\\
9     & \tabincell{c}{Number of implicit withdrawals \\with same path} & volume\\
10    & \tabincell{c}{Number of implicit withdrawals \\with different path} & volume\\
11    & Number of IGP messages & volume\\
12    & Number of EGP messages & volume\\
13    & Number of INCOMPLETE messages & volume\\
14    & Number of ORIGIN changes & volume\\
15    & Announcements to longer paths & AS-path\\
16    & Announcements to shorter paths & AS-path\\ 
17    & Average AS path length & AS-path\\ 
18    & Maximum AS path length & AS-path\\
19    & Average AS path length(unique) & AS-path\\
20    & Maximum AS path length(unique) & AS-path\\
21    & Average edit distance & AS-path\\
22    & Maximum edit distance & AS-path\\
23-33 &  \tabincell{c}{Maximum edit distance=n;
                                \\where(0,...,10)} & AS-path
                                \\
34-44 & \tabincell{c}{Maximum edit distance(unique)=n;
                                \\where(0,...,10)} & AS-path
                               \\
45    & Number of rare ASes & AS-path\\
46    & Maximum number of rare ASes & AS-path\\
\bottomrule[2pt]
\end{tabular}
\label{tab:Features_46}
\end{table}

\subsection{Experimental Setup}
Model evaluation consists of model comparison, model index calculation and model parameter selection.

\subsubsection{Comparative Method}
For comparison, we select SVM \cite{al2012machine}, NB \cite{al2012feature}, DT, RF, Ada.Boost \cite{li2014classification} methods from traditional machine learning, and these methods are still widely used in the current \cite{sanchez2019comparing, hoarau2021suitability, ding2016detecting} due to the features of high amount and high correlation. Meanwhile, the MLP \cite{teoh2018anomaly}, RNN, LSTM, MSLSTM methods from neural network models are also chosen as baselines. MSLSTM combines DWT and two-layer LSTM architectures, improves performance by focusing on the multi-time scale features of anomalous samples.

\subsubsection{Evaluation Metric}
We view the BGP anomaly detection as a classification problem, thus choose $Accuracy$, $Precision$, $Recall$, and $F1$ as the evaluation metrics. The sample dataset of real anomalous events is extremely imbalanced. For better comparison, we treat the abnormal samples as the positive class and the normal samples as the negative one. In this way, we can construct the confusion matrix as shown in Table \ref{tab:Confusion_M}.
\begin{table}[htb]
\centering
\caption{Confusion matrix}
\renewcommand\arraystretch{1.3} 
\begin{tabular}{p{3cm}<{\centering} p{2cm}<{\centering} p{2cm}<{\centering}}
\toprule[2pt]
\multirow{2}{*}{True Value} & \multicolumn{2}{c}{Predicted Value}     \\ \cline{2-3} 
                            & \multicolumn{1}{c}{Positive} & Negative \\
\midrule
Positive                    & \multicolumn{1}{c}{TP(11)}   & FN(10)   \\
Negative                    & \multicolumn{1}{c}{FP(01)}   & TN(00)   \\
\bottomrule[2pt]
\end{tabular}
\label{tab:Confusion_M}
\vspace{-3mm}
\end{table}

According to Table \ref{tab:Confusion_M}, we can calculate each performance metric as following:
\begin{small}
\begin{align}
    \text {\emph{Accuracy}}&=\frac{T P+T N}{T P+T N+F P+F N}\\
    \text {\emph{Precision}}&=\frac{T P}{T P+F P}\\
    \text {\emph{Recall}}&=\frac{T P}{T P+F N}\\
    F 1&=\frac{2 * \text {\emph{Precision}} * \text {\emph{Recall}}}{\text {\emph{Precision}}+\text {\emph{Recall}}}
\end{align}
\end{small}

\subsubsection{Parameter Setting}
The performance of a model is sensitive to the variable parameters. We split the dataset into a training set, a validation set and a testing set with a ratio of 6:1:3, with the same proportion of positive and negative samples in each subset. The training parameters are chosen as follows: learning rate $1e-4$, dropout $0.2$, and maximum epoch number $100$. This selection can make the model achieve an excellent and stable result on many datasets. Our model and all experiments are implemented on \emph{Python}, relying on the \emph{PyTorch} \cite{paszke2019pytorch} framework, the \emph{sklearn} library \cite{pedregosa2011scikit} and other related libraries and functions.

\begin{table*}[htb]
\caption{Anomaly BGP events detection results of balanced datasets}
\renewcommand\arraystretch{1.1} 
\begin{tabular}{c c c c c c c c c c c c c c}
\toprule[2pt]
\toprule
\multirow{3}{*}{Datasets}                                              & \multirow{3}{*}{\begin{tabular}[c]{@{}c@{}}Number\\ (test)\end{tabular}} & \multirow{3}{*}{\begin{tabular}[c]{@{}c@{}}Evaluation\\ indicator\end{tabular}} & \multicolumn{6}{c}{Machine Learning} & \multicolumn{4}{c}{Neural Network} & \multirow{3}{*}{\begin{tabular}[c]{@{}c@{}}Our\\ Model\end{tabular}}\\ 
\cmidrule(lr){4-9}      \cmidrule(lr){10-13}
&&& \multicolumn{1}{c}{SVM} & \multicolumn{1}{c}{NB} & \multicolumn{1}{c}{1NN} & \multicolumn{1}{c}{DT} & \multicolumn{1}{c}{RF} & \multicolumn{1}{c}{\begin{tabular}[c]{@{}c@{}}Ada\\ Boost\end{tabular}} & \multicolumn{1}{c}{MLP} & \multicolumn{1}{c}{RNN} & \multicolumn{1}{c}{LSTM} & MSLSTM \\ \midrule

\multirow{4}{*}{\begin{tabular}[c]{@{}c@{}}Code\\ Red\\ I \end{tabular}} & \multirow{4}{*}{1261} & 
Accuracy & 79.3 & 74.3 & 80.1 & 76.7 & 86.7 & \multicolumn{1}{c}{83.6} & 85.7 & 94.8 & 92.4 & \textbf{\underline{99.7}} & 98.0
\\  \cline{3-3}  & & 
Precision & 89.1 & 68.5 & 77.3 & 73.8 & 89.1 & \multicolumn{1}{c}{85.7} & 87.1 & 97.9 & 94.5 & \textbf{\underline{100.0}} & \textbf{\underline{100.0}}
\\  \cline{3-3} & &  
Recall & 57.3 & 70.7 & 73.9 & 68.2 & 77.7 & \multicolumn{1}{c}{72.6} & 77.1 & 89.8 & 87.3 & \textbf{\underline{99.4}} & 95.2
\\  \cline{3-3}  & &  
F1 & 69.8 & 69.6 & 75.6 & 70.9 & 83.0 & \multicolumn{1}{c}{78.6} & 81.8 & 93.7 & 90.7 & \textbf{\underline{99.7}} & 97.5
\\ \midrule

\multirow{4}{*}{\begin{tabular}[c]{@{}c@{}}Nimda \end{tabular}} & \multirow{4}{*}{7200} & 
Accuracy & 63.0 & 62.5 & 63.5 & 66.3 & 68.0 & \multicolumn{1}{c}{69.0} & 70.3 & 81.7 & 79.4 & \multicolumn{1}{c}{88.5} & \textbf{\underline{93.9}} 
\\  \cline{3-3}  & & 
Precision & 64.7 & 62.4 & 63.7 & 68.9 & 72.7 & \multicolumn{1}{c}{70.3} & 70.3 & 80.8 & 75.7 & \multicolumn{1}{c}{89.2} & \textbf{\underline{91.8}} 
\\  \cline{3-3}  & &  
Recall & 54.2 & 59.5 & 59.7 & 57.1 & 55.8 & \multicolumn{1}{c}{63.7} & 68.3 & 82.2 & 85.5 & \multicolumn{1}{c}{87.0} & \textbf{\underline{96.1}}
\\  \cline{3-3}  & &  
F1 & 59.0 & 60.9 & 61.6 & 62.4 & 63.1 & \multicolumn{1}{c}{66.8} & 69.3 & 81.5 & 80.3 & \multicolumn{1}{c}{88.1} & \textbf{\underline{93.9}}
\\ \midrule

\multirow{4}{*}{\begin{tabular}[c]{@{}c@{}}Slammer \end{tabular}} & \multirow{4}{*}{2881} & 
Accuracy & 68.9 & 67.8 & 70.9 & 72.1 & 69.4 & \multicolumn{1}{c}{74.6} & 78.3 & 69.9 & 66.7 & 89.7 & \textbf{\underline{98.4}} 
\\  \cline{3-3}  & & 
Precision & 77.3 & 69.7 & 76.7 & 83.4 & 91.1 & \multicolumn{1}{c}{84.8} & 87.2 & 96.6 & 58.9 & \textbf{\underline{100.0}} & 96.2
\\  \cline{3-3}  & &  
Recall & 29.3 & 31.4 & 37.0 & 35.8 & 24.3 & \multicolumn{1}{c}{42.9} & 52.4 & 24.9 & 53.0 & 74.0 & \textbf{\underline{100.0}}
\\  \cline{3-3}  & &  
F1 & 42.5 & 43.3 & 49.9 & 50.1 & 38.3 & \multicolumn{1}{c}{57.0} & 65.4 & 39.5 & 55.8 & 85.0 & \textbf{\underline{98.0}}
\\ \midrule

\multirow{4}{*}{\begin{tabular}[c]{@{}c@{}}Moscow\\ Blackout \end{tabular}} & \multirow{4}{*}{719} & 
Accuracy & 48.4 & 50.7 & 73.0 & 67.9 & 76.7 & \multicolumn{1}{c}{69.8} & 82.8 & 68.0 & 70.9 & 95.6 & \textbf{\underline{98.5}}
\\  \cline{3-3}  & & 
Precision & 33.3 & 48.7 & 66.7 & 64.1 & 71.0 & \multicolumn{1}{c}{63.7} & 80.4 & 68.4 & 68.4 & 91.9 & \textbf{\underline{100.0}}
\\  \cline{3-3}  & &  
Recall & 8.8 & 76.5 & 86.3 & 73.5 & 86.3 & \multicolumn{1}{c}{84.3} & 84.3 & 65.7 & 76.5 & \textbf{\underline{100.0}} & 96.7
\\  \cline{3-3}  & &  
F1 & 14.0 & 59.5 & 75.2 & 68.5 & 77.9 & \multicolumn{1}{c}{72.6} & 82.3 & 67.0 & 72.2 & 95.8 & \textbf{\underline{98.3}}
\\ \midrule

\multirow{4}{*}{\begin{tabular}[c]{@{}c@{}}Japan\\ Earthquake \end{tabular}} & \multirow{4}{*}{800} & 
Accuracy & 62.3 & 59.4 & 55.2 & 60.7 & 60.7 & \multicolumn{1}{c}{57.7} & 54.8 & 85.2 & 87.3 & 93.2 & \textbf{\underline{96.3}}
\\  \cline{3-3}  & & 
Precision & 63.5 & 57.0 & 53.1 & 58.5 & 59.6 & \multicolumn{1}{c}{56.2} & 52.6 & 78.0 & 83.3 & 93.7 & \textbf{\underline{100.0}}
\\  \cline{3-3}  & &  
Recall & 52.6 & 66.4 & 67.2 & 65.5 & 58.6 & \multicolumn{1}{c}{58.6} & 69.0 & \textbf{\underline{96.5}} & 92.1 & 92.1 & 92.4
\\  \cline{3-3}  & &  
F1 & 57.5 & 61.4 & 59.3 & 61.8 & 59.1 & \multicolumn{1}{c}{57.4} & 59.7 & 86.3 & 87.5 & 92.9 & \textbf{\underline{96.0}}
\\ \midrule

\multirow{4}{*}{\begin{tabular}[c]{@{}c@{}}Malaysian\\ Telecom \end{tabular}} & \multirow{4}{*}{1924} & 
Accuracy & 62.8 & 74.3 & 88.5 & 81.2 & 90.5 & \multicolumn{1}{c}{88.0} & 81.9 & 88.5 & 93.0 & 91.9 & \textbf{\underline{96.8}}
\\  \cline{3-3}  & & 
Precision & 22.8 & 53.6 & 79.9 & 61.5 & 83.0 & \multicolumn{1}{c}{77.1} & 92.9 & 74.0 & 82.6 & 82.4 & \textbf{\underline{95.9}}
\\  \cline{3-3}  & &  
Recall & 12.8 & 72.6 & 79.9 & 91.5 & 83.5 & \multicolumn{1}{c}{82.3} & 39.6 & 92.1 & \textbf{\underline{95.7}} & 91.5 & 92.2
\\  \cline{3-3}  & &  
F1 & 16.4 & 61.7 & 79.9 & 73.5 & 83.3 & \multicolumn{1}{c}{79.6} & 55.6 & 82.1 & 88.7 & 86.7 & \textbf{\underline{94.0}}
\\ \midrule \midrule

\multicolumn{2}{c}{\multirow{4}{*}{\begin{tabular}[c]{@{}c@{}}\textbf{All Dataset}\\ \textbf{Average} \end{tabular}}} & 
Accuracy & 64.1 & 64.8 & 71.9 & 70.8 & 75.3 & 73.8 & 75.6 & 81.4 & 81.6 & 93.1 & \textbf{\underline{97.0}}
\\    & & 
Precision & 58.5 & 60.0 & 69.6 & 68.4 & 77.8 & 73.0 & 78.4 & 82.6 & 77.2 & 92.9 & \textbf{\underline{97.3}}
\\    & &  
Recall & 35.8 & 62.9 & 67.3 & 65.3 & 64.4 & 67.4 & 65.1 & 75.2 & 81.7 & 90.7 & \textbf{\underline{95.4}}
\\    & &  
F1 & 43.2 & 59.4 & 66.9 & 64.5 & 67.5 & 68.7 & 69.0 & 75.0 & 79.2 & 91.4 & \textbf{\underline{96.3}}
\\

\bottomrule
\bottomrule[2pt]
\end{tabular}
\vspace*{-5mm}
\label{tab:Binary_Classification_balance}
\end{table*}

\subsection{Binary Classification}
\subsubsection{Basic Experiment}
To verify the effectiveness of our model, experiments are first conducted on a balanced sample set of six different BGP anomalous events. The details of the experiments and data are shown in Table \ref{tab:Binary_Classification_balance}. We keep the ratio of positive and negative samples of the experimental dataset to be smaller than 2:1. The duration of the Moscow and the Malaysian events is short, so multiple ASes in the event area are merged into the same dataset for training and testing. 

Table \ref{tab:Binary_Classification_balance} shows the anomaly detection results of balanced datasets for different methods. It can be seen that the evaluation indicators in our model are all above 90\% for all datasets (performs the best), and the F1 metrics are improved at most 14 percentage points compared with the baseline, which demonstrate the superiority of our model. In details, among the traditional machine learning methods, the 1NN method performs well in four classifiers (SVM, NB, 1NN, and DT), indicating that computing feature similarity on datasets which have distinct features is efficient in classification task. The results of DT, RF, and Ada.Boost show that based on ensemble learning which integrates the weak classifiers into strong classifiers is generally better than the single classifier. The performance of time series models RNN, LSTM and MSLSTM have obvious improvement on several datasets. These methods take into account the temporal dependencies in data, which are suitable for time-series datasets. Note that LSTM outperforms RNN, and the MSLSTM model with a two-layer LSTM structure performs the best.

\subsubsection{Ablation Experiment}
To study the rigor of parameter setting and the Rationality of model design, we conduct ablation experiments to answer the following questions:
\begin{itemize}
\item[$\bullet$] \textbf{Q1:} Can the period of STL and the size of the window be set arbitrarily?
\item[$\bullet$] \textbf{Q2:} Is every item decomposed by STL useful or redundant?
\item[$\bullet$] \textbf{Q3:} How does each module impact performance of our framework?
\end{itemize}

\begin{figure}[htp]
    \centering
    \includegraphics[width=8cm]{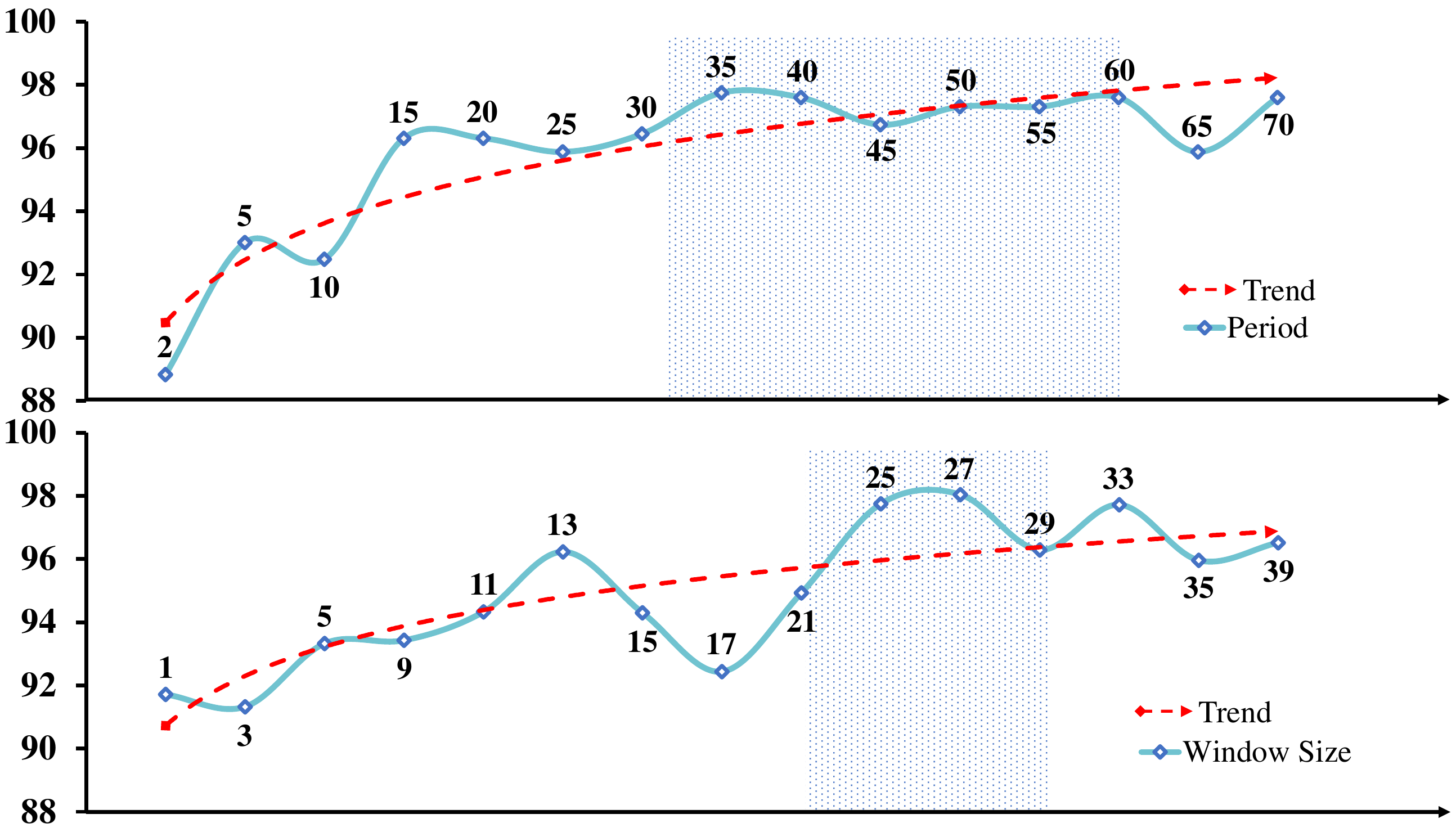}
    \caption{The parameter evaluation of window size and period.}
    \label{fig:Period_Window}
    \vspace*{-5mm}
\end{figure}

For \textbf{Q1}, it is found that the period preset value of the STL decomposition method and the size of the window slice have a significant influence on the experimental performance. Figure \ref{fig:Period_Window} presents the result of the parameter analysis experiment, where the red dashed line shows the variation trend of the model accuracy. It can be seen that as the parameter values (the period value of STL and the size of window slice) increase, the accuracy gradually reaches to a stable state. In our experiment, the window size and the period value are finally chosen to be 25 and 35(which can be fine-tuned according to the specific situation).

\begin{table}[htb]
\caption{Analysis of each decomposition term of STL.}
\centering
\renewcommand\arraystretch{1.3} 
\begin{tabular}{p{1.1cm}<{\centering} p{1.1cm}<{\centering} p{1.1cm}<{\centering} p{1.1cm}<{\centering} p{1.1cm}<{\centering} p{0.8cm}<{\centering}}
\toprule[2pt]
\multicolumn{5}{c}{Model}                  & \multirow{2}{*}{F1} 
\\ \cline{1-5}
Observed & Resid & Seasonal & Trend & Weight &                     
\\ \toprule
\checkmark & \checkmark & \checkmark & \checkmark & \checkmark & \textbf{\underline{95.88}}
\\ 
\checkmark & \checkmark & \checkmark & \checkmark & & 94.36
\\ 
\checkmark & \checkmark & \checkmark & & & 80.59
\\ 
\checkmark & \checkmark & & & & 78.65
\\ 
\checkmark & & & & & 77.34
\\
\bottomrule[2pt]
\end{tabular}
\label{tab:STL_Window}
\vspace*{-0mm}
\end{table}

\begin{figure*}[htp]
    \centering
    \includegraphics[width=18cm]{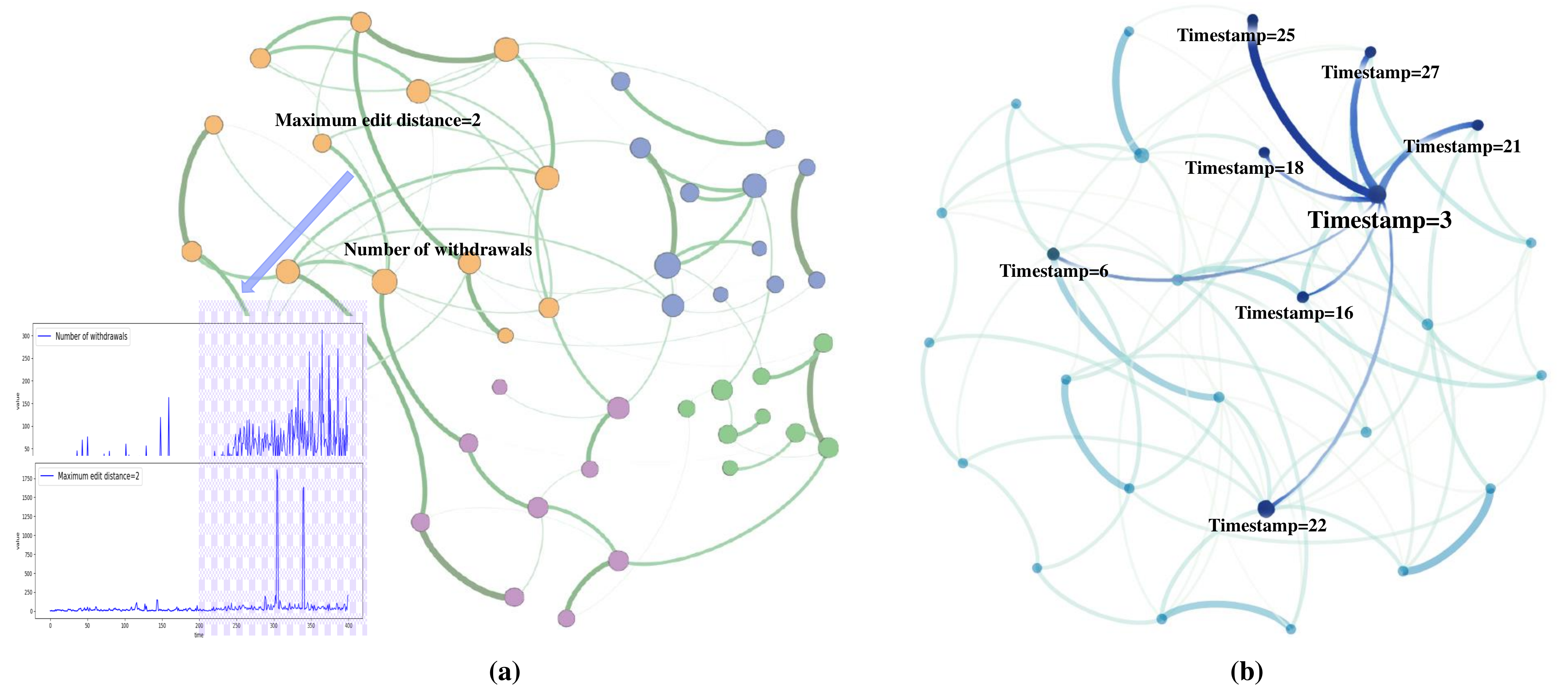}
    \caption{(a) The weighted undirected graph of the feature relationships of \emph{Nimda} uses features' attention weights as edge weights; (b) the situation of temporal dependence.}
    \label{fig:t-SNE_Interpretable}
    \vspace*{-0mm}
\end{figure*}

For \textbf{Q2}, we perform quantitative analysis of each component of the STL decomposition method in our experiment. The results are shown in Table \ref{tab:STL_Window}, which demonstrate that to achieve the optimal value of $F1$, each component is indispensable. It can be seen that, the most obvious impact on the performance is the trend term, since this term can notably reflect the difference between the positive and negative samples, as shown in Figure \ref{fig:STL_Window} (b).

For \textbf{Q3}, we verify our model by performing ablation experiment on each module. Table \ref{tab:Modules_Ablation} shows the result on the dataset Nimda, which has the most abnormal data. It can be seen that, the F1 score increases from less than 70\% to nearly 90\% through STL decomposition and window slicing, surpassing most of the baseline. It is \begin{table}[htb]
\caption{Ablation experiment of each module.}
\centering
\renewcommand\arraystretch{1.3} 
\begin{tabular}{p{0.5cm}<{\centering} p{1.2cm}<{\centering} p{1.1cm}<{\centering} p{0.8cm}<{\centering} p{1.1cm}<{\centering} p{0.6cm}<{\centering} p{0.6cm}<{\centering}}
\toprule[2pt]
\multirow{3}{*}{No.} & \multicolumn{4}{c}{Model Modules}  & \multirow{3}{*}{Acc.} & \multirow{3}{*}{F1} \\ 
\cline{2-5}
& \begin{tabular}[c]{@{}c@{}}Temporal\\ GAT\end{tabular} & \begin{tabular}[c]{@{}c@{}}Feature\\ GAT\end{tabular} & STL & \begin{tabular}[c]{@{}c@{}}Sliding\\ Window\end{tabular} 
& & \\ \toprule
1 & \checkmark & \checkmark & \checkmark & \checkmark & \textbf{\underline{93.86}} & \textbf{\underline{93.88}}
 \\ 
2 & & \checkmark & \checkmark & \checkmark & 92.64 & 92.75                    
 \\ 
3 & \checkmark & & \checkmark & \checkmark & 92.78 & 92.90    
 \\ 
4 & & & \checkmark & \checkmark & 88.60 & 89.13                     
 \\
5 & & & & \checkmark & 78.89 & 81.77                    
 \\
6 & & & & & 64.13 & 68.89 
 \\ 
\bottomrule[2pt]
\end{tabular}
\label{tab:Modules_Ablation}
\vspace*{-0mm}
\end{table}
worth mentioning that the application of the time-based and feature-based GAT can further increase the F1 score to 94\%, implying that each part of our model is indispensable.

\subsubsection{Interpretability Experiment}
\begin{figure}[htb]
    \centering
    \includegraphics[width=8cm]{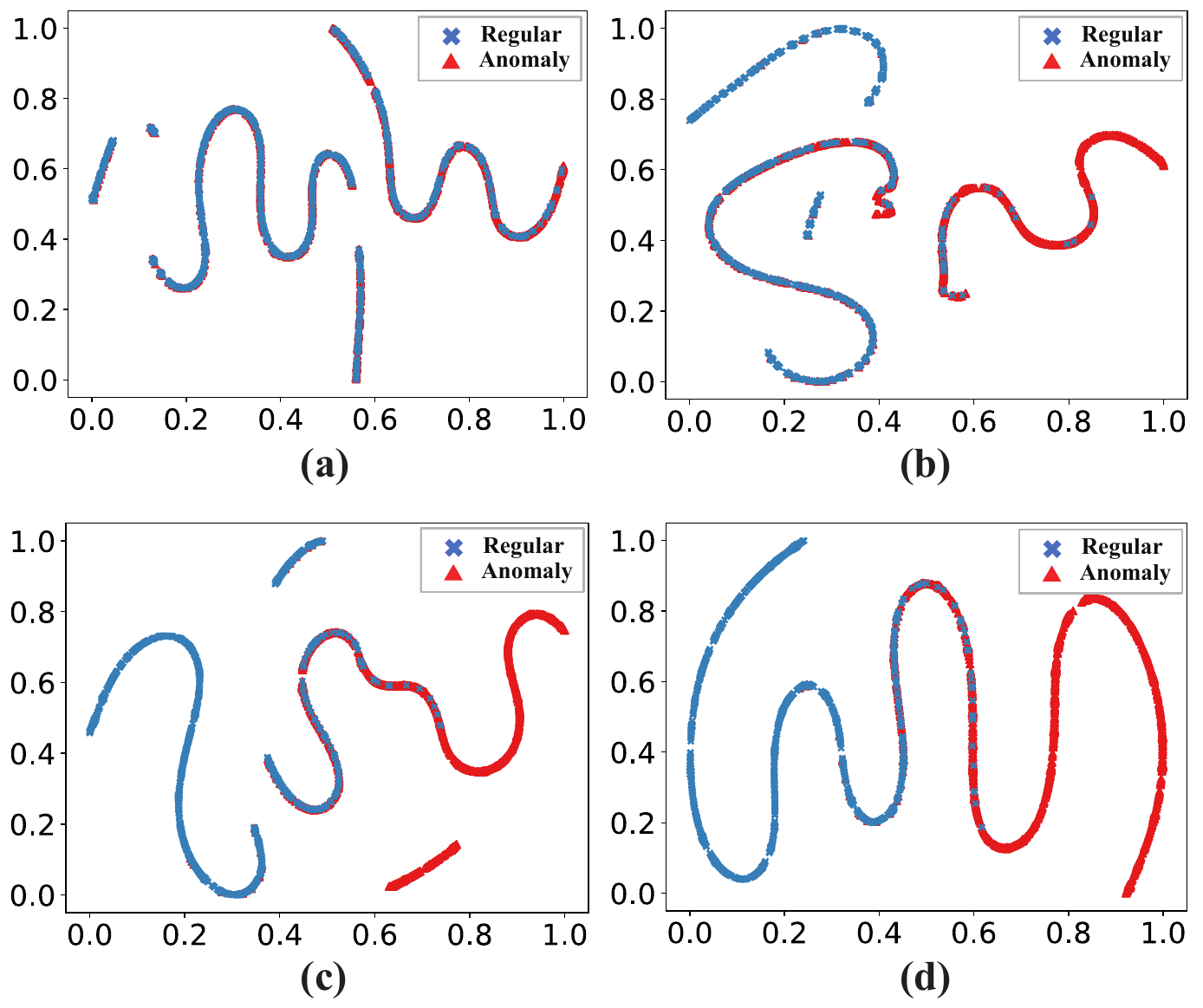}
    \caption{The t-SNE plot of the output vectors of our trained model on the \emph{Nimda} dataset.}
    \label{fig:t-SNE}
    \vspace*{-0mm}
\end{figure}

The Principle Component Analysis (PCA) \cite{abdi2010principal} dimensionality reduction method is used to visualize and analyze the output vectors in two-dimensional space. Experiments (a), (b), (c), and (d) in Figure \ref{fig:t-SNE} correspond to the experiments 6, 5, 4, and 1 in Table \ref{tab:Modules_Ablation}, respectively. It can be seen that from (a) to (d), the positive and negative samples become increasingly discriminable, which is consistent with the results as shown in Table \ref{tab:Modules_Ablation}. Figure \ref{fig:t-SNE} also demonstrates that the classification errors usually occur at the boundary of the positive and negative samples, where the data fluctuations are obvious.

In the feature-based GAT module, we construct a fully connected graph with nodes represent features and links represent the relationship between adjacent nodes. To make the visualization more significant, we plot the edges with weights great than 0.3 only and apply the community algorithm \cite{traag2019louvain} to obtain Figure \ref{fig:t-SNE_Interpretable} (a). It can be seen that in the same community, the weight between nodes is relatively large, corresponding to the strong coupling between features as observed in the time series. This strong coupling relationship of similar features is confirmed in the graph. In Figure \ref{fig:t-SNE_Interpretable} (b), the threshold value of the edge weight be set as 0.2 directly and demonstrates a time dependence on the time dimension generally favoring the medium or long term. These experiments explain the effectiveness to focus on intrinsic associations of data through GAT.

\begin{figure*}[htp]
    \centering
    \includegraphics[width=17cm]{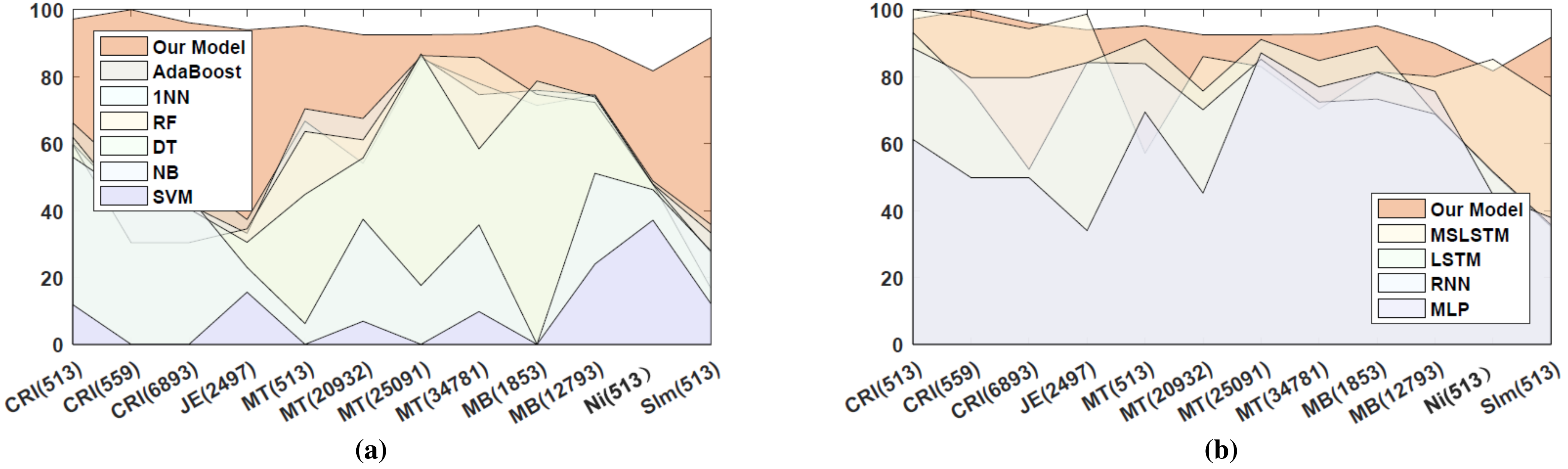}
    \caption{Comparing the F1-score of our models with machine learning models (a) and neural network models (b) on 12 imbalance datasets.}
    \label{fig:Compare}
    \vspace*{-0mm}
\end{figure*}

\subsection{Imbalance Dataset Classification}
In real situations, the dataset of the Internet events is generally imbalanced. In this section, we conduct experiments on different AS data for each event spanning five days. The maximum ratio of positive and negative samples on the test set can reach 100:1, considering that the time of some real events can only last for one or two hours. In this case, the accuracy rate can not illustrate the effectiveness of a model, one needs to consider the F1 value. Figure \ref{fig:Compare}(a) shows the F1 value on the imbalanced dataset for different methods based on machine learning. Figure \ref{fig:Compare}(b) presents the same results for different methods based on neural network model. It is shown that neural network methods outperform machine learning methods, and our model performs the best.

Table \ref{tab:Imbalance_Total} provides the detailed value of each metric on 12 datasets for all methods mentioned above. The second column specifies the total number of samples and the  number of abnormal samples contained in each event. The F1 values of our model are higher than 80\% for all datasets, even up to 99\%. Despite the good performance of MSLSTM on some datasets, means for each metric calculated from the results of 12 datasets, and our model is optimal at all metrics, and the mean of F1 metric is outstanding with nearly 10 percent higher than the best baseline. Thus the severe bias caused by sample imbalance can be mitigated effectively.

\subsection{Multi-class Detection}
Finally, we apply our model to multi-class task which is more critical in reality. We choose six  datasets from the balanced samples. For each dataset, 600 consecutive samples around the anomalous event (300 normal samples and 300 abnormal samples) are selected, forming a multi-classification dataset. The parameters are kept the same as 
before, and the comparison methods are chosen to be RNN, LSTM, and MSLSTM. The results in Table \ref{tab:Multi_Class_Balance} show that our method can increase the F1 value of positive and negative 
\begin{table}[htbp]
\centering
\caption{Results of multi-classification experiments}
\renewcommand\arraystretch{1.3} 
\begin{tabular}{c c c c c}
\toprule[2pt]
Method   & RNN & LSTM & MSLSTM & Our Model \\
\midrule
Accuracy & \textbf{\underline{93.5}}   & 88.3    & 88.8      & \textbf{\underline{93.5}} \\
F1       & 55.0   & 53.6    & 89.6      & \textbf{\underline{93.5}} \\
\bottomrule[2pt]
\end{tabular}
\label{tab:Multi_Class_Balance}
\vspace*{0mm}
\end{table}
\begin{figure}[htp]
    \centering
    \includegraphics[width=8cm]{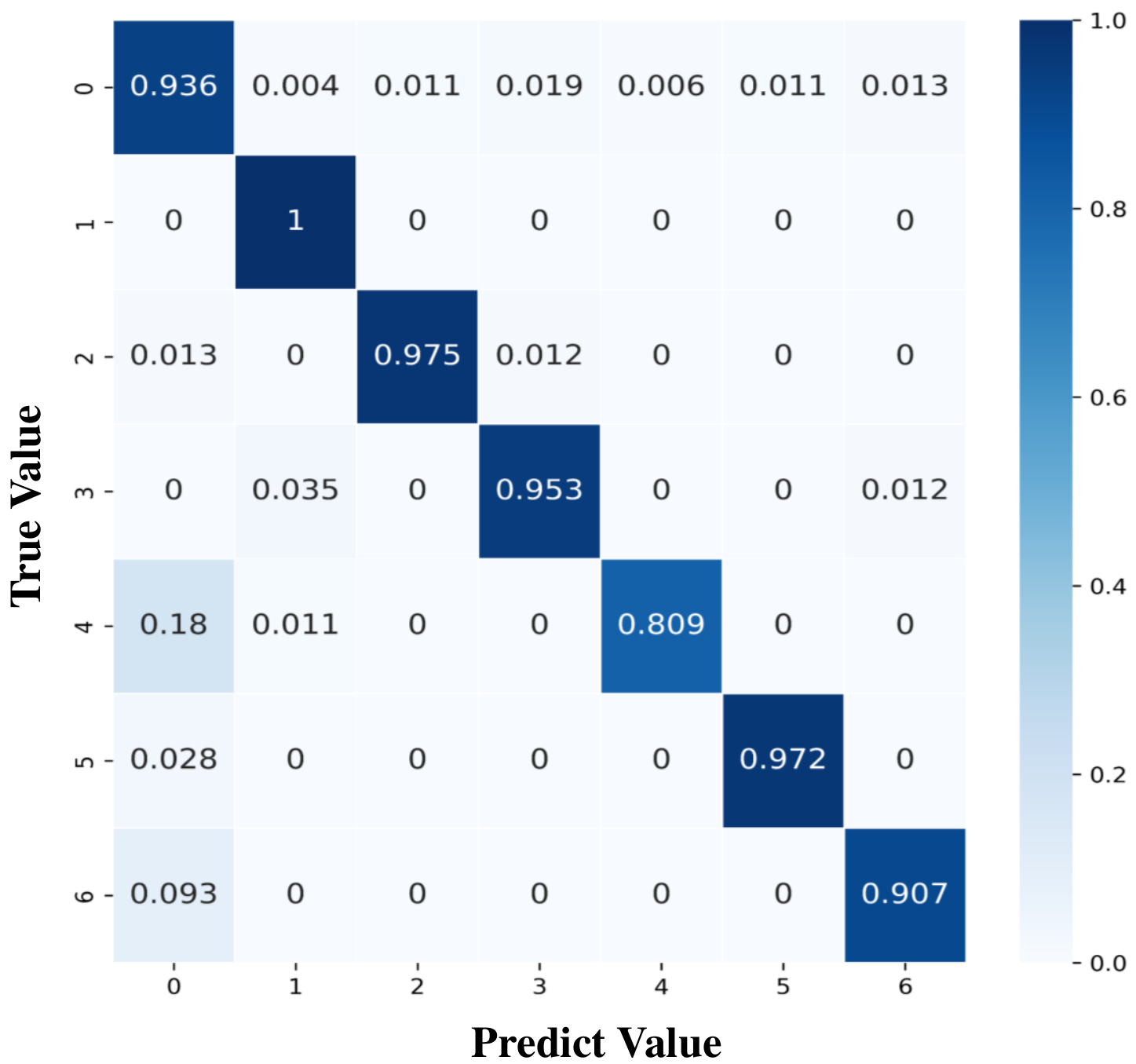}
    \caption{Heat map of multi-classification experimental results.}
    \label{fig:Multi_Class_Balance}
    \vspace*{-0mm}
\end{figure}
\begin{table*}[p]
\caption{Anomaly BGP events detection results of imbalanced datasets}
\renewcommand\arraystretch{0.95} 
\begin{tabular}{c c c c c c c c c c c c c c}
\toprule[2pt]
\toprule
\multirow{3}{*}{Datasets}                                              & \multirow{3}{*}{\begin{tabular}[c]{@{}c@{}}Number\\ (Total) \\ (Anomaly)\end{tabular}} & \multirow{3}{*}{\begin{tabular}[c]{@{}c@{}}Evaluation\\ indicator\end{tabular}} & \multicolumn{6}{c}{Machine Learning} & \multicolumn{4}{c}{Neural Network} & \multirow{3}{*}{\begin{tabular}[c]{@{}c@{}}Our\\ Model\end{tabular}}\\ 
\cmidrule(lr){4-9}      \cmidrule(lr){10-13}
&&& \multicolumn{1}{c}{SVM} & \multicolumn{1}{c}{NB} & \multicolumn{1}{c}{1NN} & \multicolumn{1}{c}{DT} & \multicolumn{1}{c}{RF} & \multicolumn{1}{c}{\begin{tabular}[c]{@{}c@{}}Ada\\ Boost\end{tabular}} & \multicolumn{1}{c}{MLP} & \multicolumn{1}{c}{RNN} & \multicolumn{1}{c}{LSTM} & MSLSTM \\ \midrule

\multirow{4}{*}{\begin{tabular}[c]{@{}c@{}}Code\\ Red\\ I \\ (513) \end{tabular}} & \multirow{4}{*}{\begin{tabular}[c]{@{}c@{}}7136\\ (526) \end{tabular}} & 
Accuracy & 93.0 & 93.7 & 95.1 & 95.1 & 95.6 & \multicolumn{1}{c}{96.1} & 95.6 & 98.5 & 99.0 & \textbf{\underline{100.0}} & 99.6
\\  \cline{3-3}  & & 
Precision & 83.3 & 57.8 & 75.0 & 73.0 & 89.7 & \multicolumn{1}{c}{90.1} & 85.2 & 97.7 & 96.6 & \textbf{\underline{100.0}} & \textbf{\underline{100.0}}
\\  \cline{3-3} & &  
Recall & 6.4 & 54.1 & 49.7 & 53.5 & 44.6 & \multicolumn{1}{c}{52.2} & 47.8 & 80.9 & 89.8 & \textbf{\underline{100.0}} & 94.6
\\  \cline{3-3}  & &  
F1 & 11.8 & 55.9 & 59.8 & 61.8 & 59.6 & \multicolumn{1}{c}{66.1} & 61.2 & 88.5 & 93.1 & \textbf{\underline{100.0}} & 97.2
\\ \midrule

\multirow{4}{*}{\begin{tabular}[c]{@{}c@{}}Code\\ Red\\ I \\ (559) \end{tabular}} & \multirow{4}{*}{\begin{tabular}[c]{@{}c@{}}7136\\ (472) \end{tabular}} & 
Accuracy & 93.4 & 93.1 & 94.3 & 93.8 & 94.4 & \multicolumn{1}{c}{95.2} & 95.2 & 97.7 & 97.4 & 99.7 & \textbf{\underline{100.0}}
\\  \cline{3-3}  & & 
Precision & 0.0 & 47.5 & 62.9 & 55.6 & 86.7 & \multicolumn{1}{c}{78.8} & 79.7 & 96.0 & 98.9 & \textbf{\underline{100.0}} & \textbf{\underline{100.0}}
\\  \cline{3-3} & &  
Recall & 0.0 & 41.1 & 31.2 & 31.9 & 18.4 & \multicolumn{1}{c}{36.9} & 36.2 & 68.1 & 61.7 & 95.7 & \textbf{\underline{100.0}}
\\  \cline{3-3}  & &  
F1 & 0.0 & 44.1 & 41.7 & 40.5 & 30.4 & \multicolumn{1}{c}{50.2} & 49.8 & 79.7 & 76.0 & 97.8 & \textbf{\underline{100.0}}
\\ \midrule

\multirow{4}{*}{\begin{tabular}[c]{@{}c@{}}Code\\ Red\\ I \\ (6893) \end{tabular}} & \multirow{4}{*}{\begin{tabular}[c]{@{}c@{}}7136\\ (526) \end{tabular}} & 
Accuracy & 93.4 & 93.1 & 94.3 & 93.8 & 94.4 & \multicolumn{1}{c}{95.2} & 95.2 & 97.7 & 95.6 & 99.2 & \textbf{\underline{99.5}}
\\  \cline{3-3}  & & 
Precision & 0.0 & 47.5 & 62.9 & 55.6 & 86.7 & \multicolumn{1}{c}{78.8} & 79.7 & 96.0 & 89.7 & \textbf{\underline{100.0}} & \textbf{\underline{100.0}}
\\  \cline{3-3} & &  
Recall & 0.0 & 41.1 & 31.2 & 31.9 & 18.4 & \multicolumn{1}{c}{36.9} & 36.2 & 68.1 & 36.9 & 89.2 & \textbf{\underline{92.5}}
\\  \cline{3-3}  & &  
F1 & 0.0 & 44.1 & 41.7 & 40.5 & 30.4 & \multicolumn{1}{c}{50.2} & 49.8 & 79.7 & 52.3 & 94.3 & \textbf{\underline{96.1}}
\\ \midrule

\multirow{4}{*}{\begin{tabular}[c]{@{}c@{}}Japan\\ Earthquake\\ (2497) \end{tabular}} & \multirow{4}{*}{\begin{tabular}[c]{@{}c@{}}7200\\ (387)  \end{tabular}} & 
Accuracy & 95.0 & 94.7 & 92.9 & 92.8 & 95.6 & \multicolumn{1}{c}{95.3} & 95.5 & 98.2 & 98.3 & \textbf{\underline{99.9}} & 99.4
\\  \cline{3-3}  & & 
Precision & 83.3 & 53.1 & 33.6 & 31.8 & 86.2 & \multicolumn{1}{c}{66.7} & 80.6 & 81.5 & 85.7 & 99.1 & \textbf{\underline{100.0}}
\\  \cline{3-3}  & &  
Recall & 8.6 & 14.7 & 32.8 & 29.3 & 21.6 & \multicolumn{1}{c}{25.9} & 21.6 & 87.1  & 82.8 & \textbf{\underline{98.3}} & 88.7
\\  \cline{3-3}  & &  
F1 & 15.6 & 23.0 & 33.2 & 30.5 & 34.5 & \multicolumn{1}{c}{37.3} & 34.0 & 84.2 & 84.2 & \textbf{\underline{98.7}} & 94.0
\\ \midrule

\multirow{4}{*}{\begin{tabular}[c]{@{}c@{}}Malaysian\\ Telecom\\ (513) \end{tabular}} & \multirow{4}{*}{\begin{tabular}[c]{@{}c@{}}7200\\ (103)  \end{tabular}} & 
Accuracy & 98.6 & 98.6 & 99.3 & 98.3 & 99.3 & \multicolumn{1}{c}{99.3} & 99.3 & 99.5 & 99.8 & 99.2 & \textbf{\underline{99.9}}
\\  \cline{3-3}  & & 
Precision & 0.0 & 50.0 & 79.2 & 40.5 & \textbf{\underline{100.0}} & \multicolumn{1}{c}{\textbf{\underline{100.0}}} & 89.5 & 81.2 & 96.3 & \textbf{\underline{100.0}} & 90.9
\\  \cline{3-3}  & &  
Recall & 0.0 & 3.3 & 63.3 & 50.0 & 46.7 & \multicolumn{1}{c}{50.0} & 56.7 & 86.7 & 86.7 & 40.0 & \textbf{\underline{100.0}}
\\  \cline{3-3}  & &  
F1 & 0.0 & 6.2 & 70.4 & 44.8 & 63.6 & \multicolumn{1}{c}{66.7} & 69.4 & 83.9 & 91.2 & 57.1 & \textbf{\underline{95.2}}
\\ \midrule

\multirow{4}{*}{\begin{tabular}[c]{@{}c@{}}Malaysian\\ Telecom\\ (20932) \end{tabular}} & \multirow{4}{*}{\begin{tabular}[c]{@{}c@{}}7200\\ (154)  \end{tabular}} & 
Accuracy & 97.5 & 97.4 & 98.8 & 98.4 & 98.7 & \multicolumn{1}{c}{98.4} & 98.4 & 98.9 & 99.2 & 99.3 & \textbf{\underline{99.8}}
\\  \cline{3-3}  & & 
Precision & 16.7 & 37.8 & 83.9 & 66.7 & 84.6 & \multicolumn{1}{c}{71.4} & 87.5 & 87.1 & \textbf{\underline{100.0}} & 79.6 & \textbf{\underline{100.0}}
\\  \cline{3-3}  & &  
Recall & 4.3 & 37.0 & 56.5 & 47.8 & 47.8 & \multicolumn{1}{c}{43.5} & 30.4 & 58.7 & 60.9 & \textbf{\underline{93.5}} & 86.1
\\  \cline{3-3}  & &  
F1 & 6.9 & 37.4 & 67.5 & 55.7 & 61.1 & \multicolumn{1}{c}{54.1} & 45.2 & 70.1 & 75.7 & 86.0 & \textbf{\underline{92.5}}
\\ \midrule

\multirow{4}{*}{\begin{tabular}[c]{@{}c@{}}Malaysian\\ Telecom \\ (25091)\end{tabular}} & \multirow{4}{*}{\begin{tabular}[c]{@{}c@{}}7200\\ (185)  \end{tabular}} & 
Accuracy & 97.0 & 94.8 & 99.3 & 99.4 & 99.4 & \multicolumn{1}{c}{99.3} & 99.4 & 99.3 & 99.5 & 99.3 & \textbf{\underline{99.7}}
\\  \cline{3-3}  & & 
Precision & 0.0 & 14.8 & 88.5 & 90.2 & 93.6 & \multicolumn{1}{c}{91.7} & 95.7 & 86.8 & 89.5 & \textbf{\underline{100.0}} & \textbf{\underline{100.0}}
\\  \cline{3-3}  & &  
Recall & 0.0 & 21.8 & 83.6 & 83.6 & 80.0 & \multicolumn{1}{c}{80.0} & 80.0 & 83.6 & \textbf{\underline{92.7}} & 70.9 & 86.0
\\  \cline{3-3}  & &  
F1 & 0.0 & 17.6 & 86.0 & 86.8 & 86.3 & \multicolumn{1}{c}{85.4} & 87.1 & 85.2 & 91.1 & 83.0 & \textbf{\underline{92.5}}
\\ \midrule

\multirow{4}{*}{\begin{tabular}[c]{@{}c@{}}Malaysian\\ Telecom\\ (34781) \end{tabular}} & \multirow{4}{*}{\begin{tabular}[c]{@{}c@{}}7200\\ (107)  \end{tabular}} & 
Accuracy & 98.3 & 96.7 & 99.2 & 98.3 & 99.6 & \multicolumn{1}{c}{99.4} & 99.4 & 99.3 & 99.5 & 98.7 & \textbf{\underline{99.9}}
\\  \cline{3-3}  & & 
Precision & 22.2 & 25.0 & 71.4 & 45.6 & \textbf{\underline{100.0}} & \multicolumn{1}{c}{85.2} & \textbf{\underline{100.0}} & 80.8 & 82.4 & 54.2 & \textbf{\underline{100.0}}
\\  \cline{3-3}  & &  
Recall & 6.2 & 62.5 & 78.1 & 81.2 & 75.0 & \multicolumn{1}{c}{71.9} & 62.5 & 65.6 & 87.5 & \textbf{\underline{100.0}} & 86.4
\\  \cline{3-3}  & &  
F1 & 9.8 & 35.7 & 74.6 & 58.4 & 85.7 & \multicolumn{1}{c}{78.0} & 76.9 & 72.4 & 84.8 & 70.3 & \textbf{\underline{92.7}}
\\ \midrule

\multirow{4}{*}{\begin{tabular}[c]{@{}c@{}}Moscow\\ Blackout \\ (1853) \end{tabular}} & \multirow{4}{*}{\begin{tabular}[c]{@{}c@{}}7200\\ (171)  \end{tabular}} & 
Accuracy & 97.6 & 97.5 & 99.0 & 99.1 & 99.0 & \multicolumn{1}{c}{98.9} & 99.2 & 98.9 & 99.5 & 99.3 & \textbf{\underline{99.8}}
\\  \cline{3-3}  & & 
Precision & 0.0 & 0.0 & 91.7 & 92.1 & 96.9 & \multicolumn{1}{c}{90.9} & 92.5 & 84.6 & \textbf{\underline{100.0}} & \textbf{\underline{100.0}} & 93.0
\\  \cline{3-3}  & &  
Recall & 0.0 & 0.0 & 64.7 & 68.6 & 60.8 & \multicolumn{1}{c}{58.8} & 72.5 & 64.7 & 80.4 & 68.6 & \textbf{\underline{97.6}}
\\  \cline{3-3}  & &  
F1 & 0.0 & 0.0 & 75.9 & 78.7 & 74.7 & \multicolumn{1}{c}{71.4} & 81.3 & 73.3 & 89.1 & 81.4 & \textbf{\underline{95.2}}
\\ \midrule

\multirow{4}{*}{\begin{tabular}[c]{@{}c@{}}Moscow\\ Blackout \\ (12793) \end{tabular}} & \multirow{4}{*}{\begin{tabular}[c]{@{}c@{}}7200\\ (171)  \end{tabular}} & 
Accuracy & 97.4 & 96.9 & 98.9 & 98.9 & 98.8 & \multicolumn{1}{c}{98.9} & 99.0 & 98.7 & 98.4 & 98.9 & \textbf{\underline{99.6}}
\\  \cline{3-3}  & & 
Precision & 37.5 & 40.7 & 81.4 & 82.9 & 79.1 & \multicolumn{1}{c}{81.4} & \textbf{\underline{87.2}} & 76.2 & 62.9 & 71.9 & 83.3
\\  \cline{3-3}  & &  
Recall & 17.6 & 68.6 & 68.6 & 66.7 & 66.7 & \multicolumn{1}{c}{68.6} & 66.7 & 62.7 & 76.5 & 90.2 & \textbf{\underline{97.6}} 
\\  \cline{3-3}  & &  
F1 & 24.0 & 51.1 & 74.5 & 73.9 & 72.3 & \multicolumn{1}{c}{74.5} & 75.6 & 68.8 & 69.0 & 80.0 & \textbf{\underline{89.9}}
\\ \midrule

\multirow{4}{*}{\begin{tabular}[c]{@{}c@{}}Nimda \\ (513) \end{tabular}} & \multirow{4}{*}{\begin{tabular}[c]{@{}c@{}}10336\\ (3535)  \end{tabular}} &
Accuracy & 67.0 & 63.7 & 66.4 & 64.7 & 69.6 & \multicolumn{1}{c}{67.3} & 67.2 & 64.2 & 69.0 & \multicolumn{1}{c}{\textbf{\underline{90.8}}} & 88.1
\\  \cline{3-3}  & & 
Precision & 53.2 & 46.9 & 51.0 & 48.4 & 58.0 & \multicolumn{1}{c}{52.5} & 52.7 & 48.0 & 55.4 & \multicolumn{1}{c}{\textbf{\underline{95.0}}} & 86.2 
\\  \cline{3-3}  & &  
Recall & 28.5 & 45.6 & 45.1 & 47.3 & 40.6 & \multicolumn{1}{c}{45.6} & 38.9 & 55.8 & 47.8 & \multicolumn{1}{c}{77.3} & \textbf{\underline{77.7}}
\\  \cline{3-3}  & &  
F1 & 37.1 & 46.2 & 47.9 & 47.8 & 47.8 & \multicolumn{1}{c}{48.8} & 44.7 & 51.6 & 51.3 & \multicolumn{1}{c}{\textbf{\underline{85.2}}} & 81.7
\\ \midrule

\multirow{4}{*}{\begin{tabular}[c]{@{}c@{}}Slammer \\ (513) \end{tabular}} & \multirow{4}{*}{\begin{tabular}[c]{@{}c@{}}7200\\ (1130)  \end{tabular}} & 
Accuracy & 83.9 & 84.0 & 86.7 & 86.3 & 85.7 & \multicolumn{1}{c}{87.4} & 87.5 & 76.4 & 84.5 & 93.5 & \textbf{\underline{97.2}} 
\\  \cline{3-3}  & & 
Precision & 41.4 & 47.2 & 76.6 & 81.2 & 93.9 & \multicolumn{1}{c}{88.4} & 86.3 & 31.2 & 50.8 & \textbf{\underline{100.0}} & 84.9
\\  \cline{3-3}  & &  
Recall & 7.1 & 19.8 & 21.3 & 16.6 & 9.2 & \multicolumn{1}{c}{22.5} & 24.3 & 41.7 & 26.9 & 58.9 & \textbf{\underline{99.7}}
\\  \cline{3-3}  & &  
F1 & 12.1 & 27.9 & 33.3 & 27.5 & 16.7 & \multicolumn{1}{c}{35.8} & 37.9 & 35.7 & 35.2 & 74.1 & \textbf{\underline{91.7}}
\\ \midrule \midrule

\multicolumn{2}{c}{\multirow{4}{*}{\begin{tabular}[c]{@{}c@{}}\textbf{All Dataset}\\ \textbf{Average} \end{tabular}}} & 
Accuracy & 92.7 & 92.0 & 93.7 & 93.2 & 94.2 & 94.2 & 94.2 & 93.9 & 95.0 & 98.2 & \textbf{\underline{98.5}} 
\\    & & 
Precision & 28.1 & 39.0 & 71.5 & 63.6 & 88.0 & 81.3 & 84.7 & 78.9 & 84.0 & 91.7 & \textbf{\underline{94.9}}
\\    & &  
Recall & 6.6 & 34.1 & 52.2 & 50.7 & 44.2 & 49.4 & 47.8 & 68.6 & 69.2 & 81.9 & \textbf{\underline{92.2}}
\\    & &  
F1 & 9.8 & 32.4 & 58.9 & 53.9 & 55.3 & 59.9 & 59.4 & 72.8 & 74.4 & 84.0 & \textbf{\underline{93.2}}
\\

\bottomrule
\bottomrule[2pt]
\end{tabular}
\vspace*{0mm}
\label{tab:Imbalance_Total}
\end{table*}
\begin{table}[htb]
\centering
\caption{Experiments of multi-class prediction.}
\renewcommand\arraystretch{1.3}
\begin{tabular}{c c c c c c}
\toprule[2pt]
\multirow{3}{*}{\begin{tabular}[c]{@{}c@{}}Train\\ Data\end{tabular}} & \multirow{3}{*}{\begin{tabular}[c]{@{}c@{}}Test\\ Data\end{tabular}} & \multicolumn{4}{c}{F1} \\ \cmidrule(lr){3-6}
 & & \multicolumn{1}{c}{\multirow{2}{*}{RNN}} & \multicolumn{1}{c}{\multirow{2}{*}{LSTM}} & \multicolumn{1}{c}{\multirow{2}{*}{MSLSTM}} & \multirow{2}{*}{\begin{tabular}[c]{@{}c@{}}Our\\ Model\end{tabular}} \\
  & & \multicolumn{1}{c}{} & \multicolumn{1}{c}{} & \multicolumn{1}{c}{} & \\ \midrule
(2,3,4,5,6) & 1 & 81.5 & 83.2 & 90.3 & \textbf{\underline{98.1}} \\ 
(1,3,4,5,6) & 2 & 55.2 & 67.5 & 70.7 & \textbf{\underline{92.2}} \\
(1,2,4,5,6) & 3 & 89.3 & 87.5 & \textbf{\underline{99.7}} & 96.2 \\
(1,2,3,5,6) & 4 & 79.5 & 81.0 & \textbf{\underline{85.7}} & 81.1 \\
(1,2,3,4,6) & 5 & 58.9 & 74.9 & 80.5 & \textbf{\underline{97.0}} \\
(1,2,3,4,5) & 6 & 58.9 & 74.9 & 80.5 & \textbf{\underline{97.0}} \\ \midrule
\multicolumn{2}{c}{\begin{tabular}[c]{@{}c@{}} \textbf{Average} \end{tabular}} & 72.6 & 76.7 & 85.7 & \textbf{\underline{93.6}} \\

\bottomrule[2pt]
\end{tabular}
\label{tab:Multi_Class_Prediction}
\vspace*{-0mm}
\end{table}
samples from 89.6\% to 93.5\%. The classification results on the test set are presented in Figure \ref{fig:Multi_Class_Balance}, which reflect that our model is applicable to a wide range of known anomalous events.

In real scenarios, however, the events for prediction are usually unknown. To explore this problem, we separately select a dataset from the whole six as the testing set with unknown states while the other five datasets constitute the training set with known states, to find anomalous samples from the testing set. From the experimental results shown in Table \ref{tab:Multi_Class_Prediction}, the average of F1 can reach 93.6\% in anomaly detection for different unknown events, which demonstrates the perfect prediction ability of our model. 

\section{Conclusion}
In this paper, we treat the BGP anomaly detection as a classification task on a multivariate time series. Considering that the raw data are generally noisy, we enhance the data quality by employing the methods like STL decomposition and window slicing. Then we apply GAT to capture the relationship between features and the temporal dependencies in data. After that, we multiply the output of the GAT layer by a weight matrix to increase the heterogeneity of different categories, and use LSTM classifier to implement anomaly classification. The experimental results show that our model has extremely high accuracy and very low false alarm rates than other baseline methods on different datasets, implying that taking into account the inherent association in data is necessary. Furthermore, we apply our model to tasks such as multi-class anomalous event detection and unknown anomalous event classification and verify the effectiveness of our model.

\section*{Acknowledgment}
This work was partially supported by the National Natural Science Foundation of China under Grant 61973273 and by the Zhejiang Provincial Natural Science Foundation of China under Grants LY21F030017 and LR19F030001. The authors would like to thank all the members of the IVSN Research Group, Zhejiang University of Technology for the valuable discussions about the ideas and technical details presented in this paper.


\bibliographystyle{model1-num-names}
\bibliography{references}



\end{document}